\documentclass{article}
\usepackage{mathtools}
\DeclarePairedDelimiter\ceil{\lceil}{\rceil}
\DeclarePairedDelimiter\floor{\lfloor}{\rfloor}
\usepackage{amsmath}
\usepackage{listings}
\usepackage{color} 
\usepackage{caption}
\usepackage{lipsum}

\definecolor{mygreen}{RGB}{28,172,0} 
\definecolor{mylilas}{RGB}{170,55,241}
\begin{document}

\centering{\bf{\large{A study of Thompson Sampling with Parameter h}}}\\
\vspace{3ex}
\centering{\large{Qiang Ha}}\\
\centering{\large{University of Cambridge}}\\
\centering{\small{qh227@cam.ac.uk}}\\
\vspace{9ex}
\centering{\bf{\small{Abstract}}}
\begin{flushleft}
\small{
\hspace{2ex}
Thompson Sampling algorithm is a well known Bayesian algorithm for solving stochastic multi-armed bandit.At each time step the algorithm chooses each arm with probability proportional to it being the current best arm.We modify the strategy by introducing a paramter h which alters the importance of the probability of an arm being the current best arm. We show that the optimality of Thompson sampling is robust to this perturbation within a range of parameter values for two arm bandits. }
\end{flushleft}
\vspace{2ex}
\begin{flushleft}
\large{\bf{1.Introduction }}\\
\end{flushleft}
\begin{flushleft}
\hspace{2ex}
A multi-armed bandit problem (MAB) is a sequential decision making problem defined by a set of actions. At each time step, an action is taken and some observable payoff is obtained. 
The goal is to maximize the cumulative payoff achieved in a fixed period of time. William R. Thompson was the first person to have studied multi-armed  bandit problems.His motivation was to figure out the best way to choose a treatment for the next patient when there were different treatments available,Many versions of MAB and its generlalistions have been studied in literature ever since.\\
\hspace{2ex}
One of the basic version of MAB is a stochastic multi-armed bandit(SMAB).In this problem, an action correspondes to choosing an arm. Once an arm is chosen, a payoff is generated by a fixed underlying distribution for that arm.We have no/little information about the distributions to start with, and we gradually learn about them as we continuously observe payoffs.A good algorithm therefore would require a balance between choosing the current best arm and gather more information about actions. \\
\hspace{2ex}
One of the earliest algorithms to solve SMAB is a Bayesian algorithm called Thompson Sampling (TS). We assume a simple prior distribution on the parameters of the reward distribution of every arm, and at any time step, play an arm according to its posterior probability of being
the best arm.Updating prior to posterior is essentially updating our information about distributions based on the payoffs we observe.\\
\hspace{2ex}
Recently, TS has attracted considerable attention mostly because it can be easily implemented.In 2012, Shipra Agrawal and Navin Goyal ,provided a logarithmic bound on expected regret of TS algorithm in time T that is close to
the optimal regret for SMAB proved by Lai and Robins.To improve our understanding of TS algorithim, in this paper, I modified the algorithm to introudce pertubation.For the first time, I have shown the optimality of Thompson sampling is robust to this perturbation within a range of parameter values for two arm bandits. This is a first step towards a deeper understanding of TS algorithm.\\
\hspace{2ex}
Before stating my results,I will and introudce notations and assumptions for proof and describe the modification I make to TS algorithm .
\end{flushleft}
\begin{flushleft}
\bf{1.1 Notation and assumptions}
\end{flushleft}
\begin{flushleft}
\hspace{2ex}
For a formal description of SMAB and TS algorithm,please refer to Shipra Agrawal and Navin Goyal,p1-3.Assuming familarity with the mathematical terms,we will mainly focus on clarifying notations.The key points to note are: in this paper,prior of all arms are assumed to be Beta(1,1) (uniform distribution) and the underlying distribution for all arms are assumed to be Bernoulli (reward is either 0 or 1). As stated by Shipra Agrawal and Navin Goyal,TS for Bernoulli bandits can be easily generalised to any stochastic bandits. Hence, the result I am going to state is general for any two arm bandit.\\
\hspace{2ex}In addition, we will assume that the first arm is the unique optimal arm($\mu_1>\mu_i$ for i$\ne$1).This makes it convienient to write the proof for the analysis. The assumption of unique optimal arm is also without loss of generality,since adding more arms with optimal mean can decrease the expected regret.It is clear that no regret will be incured in a two arm bandit if both arms are optimal.
\\
\vspace{1ex}
\hspace{2ex}
n: number of arms,j$_i$:number of plays of arm i .\\
\hspace{2ex}
s$_i$(j$_i$):number of successes of arm i among j$_i$ plays.\\
\hspace{2ex}
$\mu_i$:mean of arm i,$\hat{\theta_i(t)}$:probability of choosing arm i at time t\\
\hspace{2ex}
$\delta_i$=$\mu_1$-$\mu_i$, $\theta_i(t)$:a random draw from posterior of arm i at time t.\\
\hspace{2ex}
$R_T$: Expected cumulative regret at time T
\end{flushleft}
\begin{flushleft}
\bf{1.2 Thompson Sampling with parameter h }\\
\end{flushleft}
\begin{flushleft}
\hspace{2ex}
A parameter h is intoduced to pertube TS algorithm.The original TS algorithm samples according to the probability of an arm being the posterior best arm. Our modifcation is to choose an arm proportional to its probability of an arm being the best arm raised to the power of h.Specifically,
\[\hat{\theta_i(t)}=P(i=arg\max_{j}(\theta_j(t)))^h/\sum_{i=1}^{n} P(i=arg\max_{j}(\theta_j(t)))^h\]
This is achieved,by using exact formulas for one beta distribution draw being bigger than draws from all other distributions in practice.\\
\hspace{2ex}
The parameter h gives us control over our belief in historical information obtained.TS algorithm effectively chooses h=1.And the bigger the h, the more we trust our up-to-date information and the more likely we will exploit the current best arm. The exploitation will likely to last until either of the two things happen: \\
1.The 'best arm' after some plays turn out to be inferior than another arm, then the algorithm will switch to the arm superioir to the 'best arm'.
2.The 'best arm' is the real optimal arm. After some plays, we are increasingly sure it gives the best expected reward. However, because of the tails of beta distributions, we always have a non zero probability of exploring the other arms.
\end{flushleft}
\begin{flushleft}
\bf{1.3 Main results}
\end{flushleft}

\begin{flushleft}
\bf{Theorem 1.} 
\end{flushleft}
\begin{flushleft}
For the two-armed stochastic bandit problem (n = 2), the expected regret of Thompson Sampling with h is 
$$E[R_T]=O(logT) \hspace{1ex}when 
\begin{cases}
 \frac{1}{2} \leq h \leq min[\frac{log\frac{1-\mu_1}{\mu_1}}{log\frac{1-y}{y}},\frac{log(1-\mu_1)+ylog(\frac{\mu_1}{1-\mu_1})}{[(1-y)]log(1-y)+ylogy]}] \hspace{1ex}and\hspace{1ex} y>\frac{1}{2}\\
\frac{1}{2} \leq h \leq \frac{log(1-\mu_1)+ylog(\frac{\mu_1}{1-\mu_1})}{[(1-y)]log(1-y)+ylogy]} \hspace{1ex}and \hspace{1ex} y\hspace{1ex}\leq \frac{1}{2}\\
\end{cases}
$$
where y=$\frac{\mu_1+\mu_2}{2}$\\
\end{flushleft}
\begin{flushleft}
\bf{Remark 1}\\
\end{flushleft}
\begin{flushleft}
\hspace{2ex}The optimal lower bound for SMAB has been proved to be O(logT).TS choose h=1 and makes the lower bound sharp. My result shows that a small deviation of h value will still give a regret of optimal order.In general,too small a h would lead to too big an instantaneous regret when we have obtained good information and too big a h would lead to too big an instantaneous regret when we have obtained wrong information\footnote[1]{There are two other remarks about Theorem1 which is put at the end of the paper because they are linked closely with the proof } .\\
\end{flushleft}

\begin{flushleft}
\large{\bf{2.Proof }}\\
\end{flushleft}
\begin{flushleft}
\hspace{2ex}
We want to give a regret bound for T plays (T is large) for the two arm bandits.We assume arm 1 is the unique optimal arm.Note $\mu_1,\mu_2 \in$[0,1].Here $\delta=\mu_1-\mu_2$ Firstly,we divide T into two phases seprated by the time at which arm 2 has been played $N$ times,$N$ =$\frac{16lnT}{\delta^{2}}$.The first phase is the gathering information phase while the second phase is the exploring phase. Regret only comes from playing arm 2. Hence regret in the first phase is fixed. And the major work is bounding regret in the second phase.\\
\vspace{3ex}

Phase 1:\\
\hspace{2ex}Regret is $N*\delta$. \\
\hspace{2ex}Let $j_1$ denote the total number of times arm 1 has been played in the first phase.\\
\vspace{2ex}
Phase 2 (where most of work is done): \\
\hspace{2ex}After $N$ number of plays of arm 2, with high probability, the posterior distribution of arm 2 will be concentrated around its empirical mean and likely to be sharp.Spefically,for $j_2\geq N$,by Chenoff bound (Lemma 1)\\
\[Pr(s_2(j_2)/j_2>\mu_2+\frac{\delta}{4})=Pr(s_2(j_2)-j_2*\mu_2>j_2*\frac{\delta}{4})\leq e^{\frac{-2(j_2\frac{\delta}{4})^{2}}{j_2}}=e^{-\frac{1}{8}j_2\delta^{2} \leq \frac{1}{T^2}}\]
\end{flushleft}
\begin{flushleft}
\hspace{2ex}In the highly unlikely event $\frac{s_2(j_2)}{j_2}>\mu_2+\frac{\delta}{4}$, we bound number of play of arm 2 between two consecutive plays of arm 1 by T.So contribution to expected regret due to this highly unlikely event ever happen in phase 2 can be upper bounded by $\sum_{t=1}^{T} \delta \frac{1}{T}=\delta$\\
\hspace{2ex}Given the high probability event, $\frac{s_2(j_2)}{j_2}\leq\mu_2+\frac{\delta_2}{4}$,(we upper bound the probability of this highly likely event by 1) and we bound the expected regret in the second stage according to different number of plays of arm 1 in the first stage.\\
\hspace{2ex}In this case,every draw from arm 2 is likely to concentrate around its mean. Specifically, let y=$(\mu_1+\mu_2)/2=\mu_1-\frac{\delta}{2}$,given $\frac{s_2(j_2)}{j_2}\leq\mu_2+\frac{\delta}{4}$,at any step,$Pr(\theta_2(t)>y)=Pr(\theta_2(t)>\mu_2+\frac{\delta}{4}+\frac{\delta}{4})$\\
$\leq Pr(\theta_2>\frac{s_2(j_2)}{j_2}+\frac{\delta}{4})$\\
$=1-F^{Beta}_{s_2(j_2)+1,j_2-s_2(j_2)+1} \frac{s_2(j_2)}{j_2}+\frac{\delta}{4}$\\
$=F^{Bin}_{j_2+1,\frac{s_2(j_2)}{j_2}+\frac{\delta}{4}}(s_2(j_2))$\\
$\leq F^{Bin}_{j_2,\frac{s_2(j_2)}{j_2}+\frac{\delta}{4}}(s_2(j_2))$\\
$\leq e^{\frac{-2\delta^{2}j_2^{2}/16}{j_2}}$\\
$=T^{-2}$
\\ \footnote[2]{Conversion of $F^{Beta}$ to $F^{Bin}$ follows from Fact3.} Hence,we can lower bound the posterior probability of arm 1 bigger than arm 2.\\
$Pr(\theta_1(t)>\theta_2(t))\geq Pr(\theta_1(t)>y)Pr(\theta_2(t)<y)\geq Pr(\theta_1(t)>y)*(1-T^{-2})$\\

\hspace{2ex}Let $t_j$ be the time at which arm 1 is played the jth time. Let $Y_j$=$t_{j+1}-t_{j}-1$, i.e. the number of play of arm 2 between the j th and the $(j+1)$th play of arm 1.We want to bound $\sum_{j=j_1}^{T-1} Y_j$.\\
\hspace{2ex}Bounding distribution of $j_1$ is hard,(but is what we need to do to better understand the effect of h), for now, we can upper bound the contribution of regret from the second stage by
$\sum_{j_1=0}^{T-1} \sum_{s_j(j_1)=0}^{j_1} E[min(Y_j|s_1(j_1),T)]Pr(s_1(j_1))$(given arm 2 has good concentration as discussed above).
\end{flushleft}
\begin{flushleft}
\bf{Different cases of $j_1$ and $s_1(j_1$)}
\end{flushleft}
\begin{flushleft}
\hspace{2ex}1. $j_1$ is high,if we also have $s_1(j_1)$ is large,then $E[Y_j]$ can be easily bounded.\\
Specifically,when $j_1>\frac{16ln(T)}{\delta^2}$,and when $s_1(j_1)\geq (y+\frac{\delta}{4})j_1$,\\
$Pr(\theta_1(t)>y)$=1-$F^{Beta}_{s_1(j_1)+1,j_1-s_1(j_1)+1}(y)$\\
=$F^{Bin}_{j_1+1,y}(s_1(j_1))$\\
$\geq F^{Bin}_{j_1+1,y}(yj+\frac{\delta j_1}{4})$\\
$\geq 1-\frac{e^{4\delta/4}}{e^{2j_1\delta^{2}/16}}$\\
$\geq 1-\frac{e^{\delta}}{T^{2}}$\\
$\geq 1-\frac{3}{T^{2}}$\\
\hspace{2ex}
\footnote[3]{
The third last inequality comes from application of Chenoff holffding bound which is detailed as Lemma4.}
$Y_j$ is stochastichally dominated(in the sense of expectation) by a geometric varible with success rate $pr(\theta_1>y)pr(\theta_2<y)$,hence\\
$\hat{\theta_1(t)}=\frac{1}{1+(\frac{1}{Pr(\theta_1(t)>\theta_2(t))}-1)^{h}}$\\
$\implies E[Y_j]=\frac{1}{\hat{\theta_1(t)}}-1$\\
$=(\frac{1}{pr(\theta_1>\theta_2)}-1)^{h}$\\
$\leq (\frac{1}{(1-3T^{-2})(1-T^{-2})}-1)^{h}$\\
$\leq [8T^{-2}-6T^{-4}]^{h}$\\
\hspace{2ex}
\footnote[4]{
The last line uses $\frac{1}{1-x}\leq 1+2x$ for $0 \leq$x$\leq\frac{1}{2}$. This is applicable for our purpose because T is assumed to be very large.}
2. Given $j_1>\frac{16ln(T)}{\delta^2}$,by Chenoff bound, probability $s_1(j_1)$ takes value smaller than $(y+\frac{\delta}{4})j_1$ can be upper bounded as \\
$F^{Bin}_{j_1,\mu_1}(yj+\delta j/4)$=$F^{Bin}_{j_1,\mu_1}(\mu_1j_1-\delta j_1/4)\leq e^{-2j_1\frac{\delta^2}{16}}\leq T^{\frac{-m}{8}}=T^{-2}$\\
So the contribution to regret from large $j_1$(large and small successes in total is bounded by :\\ $\sum_{j_1=\frac{16lnT}{\delta^2}}^{T} \sum_{s_1(j_1)=0}^{j_1} E[Y_{j_1}|s_1(j_1)]Pr(s_1(j_1))$\\
$\leq T[(1-T^-2) [8T^{-2}-6T^{-4}]^{h}+T^{-1}]$\\
$\leq 8T^{-2h+1}+1$\\
\begin{flushleft}
\it{When h$\geq \frac{1}{2}$,the contribution to regret for large $j_1$ total is at most a constant.}\\
\end{flushleft}
\hspace{2ex}3.When $j_1$ is small,the argument is more delicate.\\
\hspace{2ex}For $s_1(j_1)\geq \ceil*{y(j_1+1)}$,
$Pr(\theta_1>y)\leq 1/2$ by fact 2.\\
$\implies \sum_{s_1(j_1)=\ceil*{y(j_1+1)}}^{j_1}E[y_j|s_1(j_1)]Pr(s_1(j_1))$
$\leq\sum_{s_1(j_1)=\ceil*{y(j_1+1)}}^{j_1} f^{Bin}_{j_1,\mu_1} s_1(j_1)*(\frac{2}{1-T^{-2}}-1)^{h}$
$\leq (\frac{2}{1-T^{-2}}-1)^{h}$\\
$=1+2hT^{-2}+O(T^{-4})$\\
\hspace{2ex}For $j_1\leq \frac{mln(T)}{\delta^2}$,$s_{1}(j_1)< \floor*{y(j+1)}$,we use tight estimate for binomial to bound regret.We let $R$=$\frac{\mu_1(1-y)^{h}}{y^h  (1-\mu_1)}$. Notice when h=1,R $>$1.\\
$\sum_{s_1(j_1)=0}^{\floor{yj_{1}}}E[Y_{j_1}|s_1(j_1)]Pr(s_1(j_1))\leq \sum_{s_1(j_1)=0}^{\floor{yj_{1}}}f_{j_1,\mu}^{Bin}(s_1(j_1)) (\frac{1}{(1-T^{-2})F^{Bin}_{j_1+1,y} s_1(j_1)}-1)^{h}$\\
$\leq \sum_{s_1(j_1)=0}^{\floor{yj_{1}}} f_{j_1,\mu}^{Bin} s_{1}(j_1) (\frac{1}{(1-T^{-2})(1-y)F^{Bin}_{j_1,y} s_1(j_1)})^{h}$\\
\begin{equation}\label{a}
\leq \sum_{s_1(j_1)=0}^{\floor{yj_{1}}} (\frac{1}{(1-T^{-2})(1-y)})^{h} f_{j_1,\mu}^{Bin} s_{1}(j_1) (\frac{1}{f^{Bin}_{j_1,y} s_1(j_1)})^{h}
\end{equation}
$
= \sum_{s_1(j_1)=0}^{\floor{yj_{1}}} (\frac{1}{(1-T^{-2})(1-y)})^{h}\frac{\mu_1^{s_1(j_1)}(1-\mu_1)^{j_1-s_1(j_1)}}{(y^h)^{s_1(j_1)}((1-y)^h)^{j_1-s_1(j_1)}} (\frac{1}{\binom{j_1}{s_1(j_1)}})^{h-1}$\\
\begin{equation}\label{b}
\leq \sum_{s_1(j_1)=0}^{\floor{yj_{1}}} (\frac{1}{(1-T^{-2})(1-y)})^{h} R^{s_1(j_1)} \frac{(1-\mu_1)^{j_1}}{(1-y)^{h(j_1)}}(\frac{1}{\binom{j_1}{s_1(j_1)}})^{h-1}
\end{equation}
In the second line,we used $F^{Bin}_{j+1,y}(a)=y F^{Bin}_{j,y}(a-1)+(1-y)F^{Bin}_{j,y}(a)\geq (1-y)F^{Bin}_{j_1,y} a$ and we have thrown away -1 because $x^a$ is an increasing function for non negative x and positive a\\
\vspace{1ex}
\hspace{2ex}To further proceed from equation \ref{b},we divide into three cases by considering the relative size of h and 1.\\
\hspace{2ex}For h=1,$R=\frac{\mu_1(1-y)}{y (1-\mu_1)}$,R $>$1.Proceeding from equation \ref{b},we have\\
$\sum_{s_1(j_1)=0}^{\floor{yj_{1}}}E[Y_{j_1}|s_1(j_1)]Pr(s_1(j_1))$\\
$\leq \sum_{s_1(j_1)=0}^{\floor{yj_{1}}} (\frac{1}{(1-T^{-2})(1-y)}) R^{s_1(j_1)} \frac{(1-\mu_1)^{j_1}}{(1-y)^{(j_1)}}$\\
$\leq \frac{1}{(1-T^{-2})(1-y)} \frac{R^{\floor{yj_1}+1}-1}{R-1}\frac{(1-\mu_1)^{j_1}}{(1-y)^{(j_1)}}$\\
$\leq \frac{1}{(1-T^{-2})(1-y)} \frac{R}{R-1}\frac{\mu_1^{yj_1}(1-\mu_1)^{j_1-yj_1}}{y^{yj_1}(1-y)^{j-yj}}$\\
$= \frac{1}{(1-T^{-2})} \frac{\mu_1}{\mu_1-y}e^{-D(y||\mu_1)j_1}$\\

\hspace{2ex}If $\floor{yj_1}< \ceil{yj_1}< \ceil{y(j_1+1)}$,then we need to consider $s_1(j_1)$=$\ceil{yj_1}$.For $s_1(j_1)=\ceil{yj_1}$,
$E[Y_{j_1}|s_1(j_1)]\leq f_{j_1,\mu}^{Bin}(s_1(j_1)) (\frac{1}{(1-T^{-2})F^{Bin}_{j_1+1,y} s_1(j_1)}-1)$\\
$\leq  f_{j_1,\mu}^{Bin} s_{1}(j_1) (\frac{1}{(1-T^{-2})(1-y)F^{Bin}_{j_1,y} s_1(j_1)})$\\
$\leq f_{j_1,\mu}^{Bin} s_{1}(j_1) (\frac{1}{(1-T^{-2})(1-y)f^{Bin}_{j_1,y} s_1(j_1)})$\\
$
= (\frac{1}{(1-T^{-2})(1-y)})\frac{\mu_1^{s_1(j_1)}(1-\mu_1)^{j_1-s_1(j_1)}}{(y^h)^{s_1(j_1)}((1-y)^h)^{j_1-s_1(j_1)}} $\\
$\leq (\frac{1}{(1-T^{-2})(1-y)}) R^{s_1(j_1)} \frac{(1-\mu_1)^{j_1}}{(1-y)^{j_1}}$\\
$\leq (\frac{1}{(1-T^{-2})(1-y)}) R^{yj_1+1} \frac{(1-\mu_1)^{j_1}}{(1-y)^{j_1}}  \text{becuase $\ceil{yj_1}\leq yj_1+1$}$ \\
$\leq (\frac{R}{(1-T^{-2})(1-y)}) e^{-D(y||\mu_1)j_1} $\\

\hspace{2ex}Therefore,for h=1\\
$$\sum_{j_1=0}^{16lnT/\delta^{2}}\sum_{s_1(j_1)=0}^{yj_1}  E[Y_{j_1}|s_1(j_1)]Pr(s_1(j_1))$$\\
$$\leq \sum_{j_1=0}^{16lnT/\delta^{2}} [(\frac{R}{(1-T^{-2})(1-y)}) e^{-D(y||\mu_1)j_1}+\frac{1}{(1-T^{-2})} \frac{\mu_1}{\mu_1-y}e^{-D(y||\mu_1)j_1}] $$\\
$$= [(\frac{R}{(1-T^{-2})(1-y)})+\frac{1}{(1-T^{-2})} \frac{\mu_1}{\mu_1-y}] \frac{1}{1-e^{-D(y||\mu_1)}} $$\\
\hspace{2ex}So the order of contribution to expected regret from this part for h=1 case is at most constant.\\
\vspace{3ex}
\hspace{2ex}For h$<$1,from equation \ref{a},the smaller the h, the smaller $(\frac{1}{f^{Bin}_{j_1,y} s_1(j_1)})^{h}$ term, hence the smaller the contribution to regret(statememt also true for the additional $s_1(j_1)=\ceil{yj_1}$ case).\\
\hspace{2ex}Hence,for h$<$1,the contribution to regeret from this part is no bigger than the h=1 case. \\
$$\sum_{j_1=0}^{16lnT/\delta^{2}}\sum_{s_1(j_1)=0}^{yj_1}  E[Y_{j_1}|s_1(j_1)]Pr(s_1(j_1))$$\\
$$\leq [(\frac{R}{(1-T^{-2})(1-y)})+\frac{1}{(1-T^{-2})} \frac{\mu_1}{\mu_1-y}] \frac{1}{1-e^{-D(y||\mu_1)}} $$\\
\hspace{2ex}So the order of contribution to expected regret for h=1 case is at most constant.\\
\vspace{3ex}
\hspace{2ex}For $h>1$, to proceed from equation \ref{b},we upper bound $(\frac{1}{\binom{j_1}{s_1(j_1)}})^{h-1}$ by 1.\\
$\sum_{s_1(j_1)=0}^{\floor{yj_{1}}}E[Y_{j_1}|s_1(j_1)]Pr(s_1(j_1))$\\
$\leq \sum_{s_1(j_1)=0}^{\floor{yj_{1}}} (\frac{1}{(1-T^{-2})(1-y)})^{h} R^{s_1(j_1)} \frac{(1-\mu_1)^{j_1}}{(1-y)^{h(j_1)}}$\\
$\leq (\frac{1}{(1-T^{-2})(1-y)})^{h} \frac{R^{\floor{yj_1}+1}-1}{R-1} \frac{(1-\mu_1)^{j_1}}{(1-y)^{h(j_1)}}$\\

\hspace{2ex}When $R>1$,we upper bound it by $\frac{R^{yj_1+1}}{R-1}$ otherwise we upper bound it by $\frac{1}{1-R}$.Hence, $$\sum_{s_1(j_1)=0}^{\floor{yj_{1}}}E[Y_{j_1}|s_1(j_1)]Pr(s_1(j_1))\leq 
\begin{cases}
(\frac{1}{(1-T^{-2})(1-y)})^{h}  (\frac{(1-\mu_1)}{(1-y)^{h}})^{j_1} \frac{1}{1-R},\text{for h$\geq$1,and R$<$1}\\ 
(\frac{1}{(1-T^{-2})(1-y)})^{h} (\frac{R^y(1-\mu_1)}{(1-y)^{h}})^{j_1} \frac{R}{R-1},\text{for h$\geq$1,and R$\geq$1}
\end{cases}
$$\\

\hspace{2ex}As before,if $\floor{yj_1}< \ceil{yj_1}< \ceil{y(j_1+1)}$,then we need to consider s=$\ceil{yj_1}$.For $s_1(j_1)=\ceil{yj_1}$,\\
$E[Y_{j_1}|s_1(j_1)]$\\
$\leq f_{j_1,\mu}^{Bin}(s_1(j_1)) (\frac{1}{(1-T^{-2})F^{Bin}_{j_1+1,y} s_1(j_1)}-1)^{h}$\\
$\leq f_{j_1,\mu}^{Bin} s_{1}(j_1) (\frac{1}{(1-T^{-2})(1-y)f^{Bin}_{j_1,y} s_1(j_1)})^{h}$\\
$\leq (\frac{1}{(1-T^{-2})(1-y)})^{h} R^{yj_1+1} \frac{(1-\mu_1)^{j_1}}{(1-y)^{h(j_1)}}(\frac{1}{\binom{j_1}{s_1(j_1)}})^{h-1}  $ \\
$\leq (\frac{1}{(1-T^{-2})(1-y)})^{h}  (\frac{R^y(1-\mu_1)}{(1-y)^{h}})^{j_1} R$\\
\hspace{2ex} Let S=$\frac{(1-\mu_1)}{(1-y)^{h}}$ ,U=$R^yS$\\
\hspace{2ex} For h$>$1,R$<$1(See Lemma 5 for conditions),we have $U<S$ and \\
$\sum_{j_1=0}^{16lnT/\delta^{2}}\sum_{s_1(j_1)=0}^{yj_1}  E[Y_{j_1}|s_1(j_1)]Pr(s_1(j_1))$\\
$\leq \sum_{j_1=0}^{16lnT/\delta^{2}} [(\frac{1}{(1-T^{-2})(1-y)})^{h}  (\frac{(1-\mu_1)}{(1-y)^{h}})^{j_1} \frac{1}{1-R}+(\frac{1}{(1-T^{-2})(1-y)})^{h}  (\frac{R^y(1-\mu_1)}{(1-y)^{h}})^{j_1} R] $\\
$\leq (\frac{1}{(1-T^{-2})(1-y)})^{h}  [\frac{1}{1-R} \frac{S^{16lnT/\delta^{2}+1}-1}{S-1}+  \frac{U^{16lnT/\delta^{2}+1}-1}{U-1}R] $\\
$\leq (\frac{1}{(1-T^{-2})(1-y)})^{h}  [\frac{1}{1-R}+R] \frac{S^{16lnT/\delta^{2}+1}-1}{S-1} $\\
$\leq (\frac{1}{(1-T^{-2})(1-y)})^{h}  [\frac{1}{1-R}+R] \frac{S}{S-1} S^{16lnT/\delta^{2}}$\\
$\leq (\frac{1}{(1-T^{-2})(1-y)})^{h}  [\frac{1}{1-R}+R] \frac{S}{S-1} T^{16lnS/\delta^{2}}$\\
\vspace{3ex}
\hspace{2ex} And for h$>$1,R$\geq$1,(See Lemma 5 for conditions),we have $U\geq S$,\\
$\sum_{j_1=0}^{16lnT/\delta^{2}}\sum_{s_1(j_1)=0}^{yj_1}  E[Y_{j_1}|s_1(j_1)]Pr(s_1(j_1))$\\
$\leq \sum_{j_1=0}^{16lnT/\delta^{2}} [(\frac{1}{(1-T^{-2})(1-y)})^{h} (\frac{R^y(1-\mu_1)}{(1-y)^{h}})^{j_1} \frac{R}{R-1}+(\frac{1}{(1-T^{-2})(1-y)})^{h}  (\frac{R^y(1-\mu_1)}{(1-y)^{h}})^{j_1} R] $\\
$\leq (\frac{1}{(1-T^{-2})(1-y)})^{h}  [\frac{R}{R-1}+R] \frac{U^{16lnT/\delta^{2}+1}-1}{U-1} $\\
$\leq (\frac{1}{(1-T^{-2})(1-y)})^{h}  [\frac{R}{R-1}+R] \frac{U}{U-1} T^{16lnU/\delta^{2}}$\\
\vspace{1ex}
\hspace{2ex}
And finally we put expected regret from all parts together to provide a final regret bound.\\
\vspace{3ex}
\begin{flushleft}
\bf{Final regret bound}
\end{flushleft}
\hspace{2ex}When h$\leq$1,\\
$\sum_{t=1}^{T} R_t$\\
$\leq \frac{16lnT}{\delta}+\delta+(8T^{-2h+1}+1)\delta+\sum_{j_1=0}^{\frac{16lnT}{\delta^2}}[1+2hT^{-2}+O(T^{-4})]\delta+O(1)\delta $\\
$\leq \frac{32lnT}{\delta}+2\delta+8T^{-2h+1}\delta+O(1)\delta $\\
\hspace{2ex}Hence for $\frac{1}{2}\leq h \leq 1$,the overall expected regret is of O(lnT). And for $0\leq h<\frac{1}{2}$, the overall expected regret is of $O(T^{-2h+1})$\\
\vspace{3ex}
\hspace{2ex}When h$>$1,and $y>\frac{1}{2}$ and $h>\frac{log\frac{1-\mu_1}{\mu_1}}{log\frac{1-y}{y}}$ (R$<$1),we have $S>1$ (see Lemma 6)\\
$\sum_{t=1}^{T} R_t$\\
$\leq \frac{16lnT}{\delta}+\delta+(8T^{-2h+1}+1)\delta+\sum_{j_1=0}^{\frac{16lnT}{\delta^2}}[1+2hT^{-2}+O(T^{-4})]\delta+(\frac{1}{(1-T^{-2})(1-y)})^{h}  [\frac{1}{1-R}+R] \frac{S}{S-1} T^{16lnS/\delta^{2}}\delta $\\
$\leq \frac{32lnT}{\delta}+2\delta+8T^{-2h+1}\delta+ \delta O(T^{16lnS/\delta^{2}})$\\
\hspace{2ex}Therefore, when S$< e^{\delta^2/16}$(iff $h< \frac{log(\frac{1-\mu_1}{e^{\delta^2/16}})}{log(1-y)}$),the overall regret is of order  $O(T^{16lnS/\delta^{2}})$.When S$\geq e^{\delta^2/16}$ $(iff h\geq \frac{log(\frac{1-\mu_1}{e^{\delta^2/16}})}{log(1-y)}$),the best we can say is that expected regret is bounded by T .\\
\vspace{3ex}
\hspace{2ex}When h$>$1,and $y\leq\frac{1}{2}$ or $h\leq \frac{log\frac{1-\mu_1}{\mu_1}}{log\frac{1-y}{y}}$ (R$\geq$1)\\
$\sum_{t=1}^{T} R_t$\\
$\leq \frac{16lnT}{\delta}+\delta+(8T^{-2h+1}+1)\delta+\sum_{j_1=0}^{\frac{16lnT}{\delta^2}}[1+2hT^{-2}+O(T^{-4})]\delta+(\frac{1}{(1-T^{-2})(1-y)})^{h}  [\frac{R}{R-1}+R] \frac{U}{U-1} T^{16lnU/\delta^{2}}\delta $\\
$\leq \frac{32lnT}{\delta}+2\delta+8T^{-2h+1}\delta+ \delta O(T^{16lnU/\delta^{2}})$\\
\hspace{2ex} We have further two cases.When $y>\frac{1}{2} \hspace{1ex}and \hspace{1ex}h\leq min[\frac{log\frac{1-\mu_1}{\mu_1}}{log\frac{1-y}{y}},\frac{log(1-\mu_1)+ylog(\frac{\mu_1}{1-\mu_1})}{[(1-y)]log(1-y)+ylogy]}]$, we have U$\leq$1. (See Lemma 7) and the overall regret is of order logT.\\
\hspace{2ex}Also,when $y\leq \frac{1}{2}$,we have, when U$\leq$1(iff $h\leq\frac{log((1-\mu_1)R^y)}{log(1-y)}$), the overall regret is of order logT.And when  1$<$U$\leq e^{\delta^2/16}$(iff $\frac{log((1-\mu_1)R^y)}{log(1-y)}<h< \frac{log(\frac{(1-\mu_1)R^y}{e^{\delta^2/16}})}{log(1-y)}$),the overall regret is of order  $O(T^{16lnU/\delta^{2}})$.When U$\geq e^{\delta^2/16}$(iff $h\geq \frac{log(\frac{(1-\mu_1)R^y}{e^{\delta^2/16}})}{log(1-y)}$),the best we can say is that expected regret is bounded by T .\\

\vspace{3ex}

\end{flushleft}
\begin{flushleft}
\large{\bf{Further remark about results}}\\
\end{flushleft}
\begin{flushleft}
\hspace{2ex}1.The importance of size of y is that when $j_1$ and $s_1(j_1)$ are both very small, the bigger y,the worse information obtained and hence for bigger y,we require a smaller h to ensure overall logarithmic order regret.\\
\hspace{2ex}2.Also,the results in the paper are obtained without considering the likelihood of number of plays $j_1$ in the first stage. We upper bounded $Pr(j_1=a)$ by 1 for all $a\leq \frac{mlnT}{\delta^2}$ .To better understand the effect of h,need to consider the probability distribution of $j_1$(the number of plays of arm 1 in the first satge). This is because number of plays of $j_1$ influences the likelihood of good/bad information.\\
\vspace{3ex}

\end{flushleft}
\begin{flushleft}
\large{\bf{Facts and theorems used in analysis}}\\
\end{flushleft}
\vspace{3ex}
\begin{flushleft}
\bf{Lemma1 Chenoff -Hoeffding bounds}\\
\end{flushleft}
\begin{flushleft}
Let $X_1,...X_n$ be random varibles with common range [0,1],and such that $E[X_1|X_2,..X_n]=\mu$. Let $S_n$=$X_1+...+X_n$,Then for all $a\geq 0$\\
$Pr(S_n\geq n\mu+a)\leq e^{-2a^{2}/n}$,
$Pr(S_n\leq n\mu-a)\leq e^{-2a^{2}/n}$.\\
\end{flushleft}
\begin{flushleft}
\bf{Fact2 Median of binomial}\\
\end{flushleft}
\begin{flushleft}
Median of the binomial distribution Bin(n,p) is either  $\floor*{np}$ or $\ceil*{np}$
\end{flushleft}
\begin{flushleft}
\bf{Lemma 3 Converting Beta to Binomial}\\
\end{flushleft}
\begin{flushleft}
$F^{Beta}_{\alpha,\beta}(y)$=1-$F^{Bin}_{\alpha+\beta-1,y}(\alpha-1)$ for $\alpha$,$\beta$ non negative integers\\
The idea of the proof is that order statistics of uniform i.i.d variables have a Beta distribution.\\
\end{flushleft}
\begin{flushleft}
\bf{Lemma4}:\\
\end{flushleft}
\begin{flushleft}
For all n,p$\in[0,1]$,$\Delta\geq 0$,$F^{Bin}_{n+1,p}(np+n\Delta)\geq 1-\frac{e^{4\Delta}}{e^{2n\Delta^2}}$.\\
Proof:
$F^{Bin}_{n+1,p}(np+n\Delta)=(1-p)F^{Bin}_{n,p}(np+n\Delta)+pF^{Bin}_{n,p}(np+n\Delta-1)\geq F^{Bin}_{n,p}(np+n\Delta-1)$\\
By Chernoff-hoeffding bounds,\\
$1-F^{Bin}_{n,p}(np+n\Delta-1)\leq e^{-2(\Delta n-1)^{2}/n}=e^{-2(n^2\Delta^2+1-2\Delta n)/n}\leq e^{-2n\Delta^2+4\Delta}=\frac{e^{4\Delta}}{e^{2n\Delta^{2}}}$\\
Hence the result in the lemma follows.\\
\end{flushleft}
\begin{flushleft}
\bf{Lemma5}:\\
\end{flushleft}
\begin{flushleft}
Let $\mu_1>y$ as defined for the two arm case, note $\frac{1-y}{y}>\frac{1-\mu_1}{\mu_1}$,let R=$\frac{\mu_1(1-y)^h}{y^h(1-\mu_1)}$,then given h$>$1,\\
R$<$1 iff $y>\frac{1}{2}$ and $h>\frac{log\frac{1-\mu_1}{\mu_1}}{log\frac{1-y}{y}}$\\
R$\geq 1$ iff $y\leq\frac{1}{2}$ or $h\leq\frac{log\frac{1-\mu_1}{\mu_1}}{log\frac{1-y}{y}}$\\
proof:
\begin{equation}\label{c}
R<1\implies \frac{1-\mu_1}{\mu_1}>(\frac{(1-y)}{y})^{h}
\end{equation}
When $y>\frac{1}{2}$,we have $\frac{1-y}{y}<1$ and so $\frac{1-\mu_1}{\mu_1}<1$\\
Therefore,equation \ref{c} beocmes $h>\frac{log\frac{1-\mu_1}{\mu_1}}{log\frac{1-y}{y}}$\\
When $y\leq \frac{1}{2}$,,we have $\frac{1-y}{y}\geq 1$,equation \ref{c} has no solution unless we also have $\frac{1-\mu_1}{\mu_1}\geq 1$,(which requires $\mu_1\leq \frac{1}{2}$).\\
In which case equation \ref{c} becomes  $h<\frac{log\frac{1-\mu_1}{\mu_1}}{log\frac{1-y}{y}}<1$,contradiction\\
Hence,$R<1 \implies y>\frac{1}{2}$ and $h>\frac{log\frac{1-\mu_1}{\mu_1}}{log\frac{1-y}{y}}$,all above is clearly reversible so this is an iff statement.
\end{flushleft}
\begin{flushleft}
\bf{Lemma6}:\\
\end{flushleft}
\begin{flushleft}
When h$>$1,and $y>\frac{1}{2}$ and $h>\frac{log\frac{1-\mu_1}{\mu_1}}{log\frac{1-y}{y}}$ (R$<$1),we have $S>1$.\\
Proof:\\
$S>1$ iff $h>\frac{log(1-\mu_1)}{log(1-y)}$\\
Also,\\
$\frac{log\frac{1-\mu_1}{\mu_1}}{log\frac{1-y}{y}}-\frac{log(1-\mu_1)}{log(1-y)}$\\
$=\frac{-log(\mu_1)log(1-y)+log(y)log(1-\mu_1)}{(log(1-y)-log(y))(log(1-y))}$\\
$>0$ \text{using conditions on y and $\mu_1$}\\
Hence the result.
\end{flushleft}
\begin{flushleft}
\bf{Lemma7}:\\
\end{flushleft}
\begin{flushleft}
When h$>$1,and $R<1$,$h\leq \frac{log(1-\mu_1)+ylog(\frac{\mu_1}{1-\mu_1})}{[(1-y)]log(1-y)+ylogy]}$, we have U$\leq$1. \\
Proof:\\
$U\leq1$ $\iff h \leq \frac{log((1-\mu_1)R^y)}{log(1-y)}$\\
 $\iff log(1-y)h \geq log(1-\mu_1)+ylogR$\\
 $\iff h[(1-y)]log(1-y)+ylogy]\geq log(1-\mu_1)+ylog(\frac{\mu_1}{1-\mu_1})$\\
$\iff h\leq \frac{log(1-\mu_1)+ylog(\frac{\mu_1}{1-\mu_1})}{[(1-y)]log(1-y)+ylogy]}$\\

$R<1$ is to ensure $h>1$ is not violated.
\end{flushleft}

\newpage
\centering{\bf{Reference:}}
\begin{flushleft}
\text{Shipra and NavinGo . Analysis of Thompson Sampling for the Multi-armed Bnadit Problem}\\\text{JMLR:Workshop and Conference Proceedings vol 23 (2012) 39.1-39.26}\\
\vspace{3ex}
\text{S.Bubeck and N. Cesa,Regret Analysis of Stochastic and Nonstochastic Multi-armed Bandit Problems}\\
\text{Foundations and Trends® in Machine Learning Proceedings Volume 5 (2012) 1-122}\\
\vspace{3ex}
\text{O.-A Maillard,R.Munos,and G.Stoltz. Finite time analysis of multi-armed bandit problems}\\
\text{with kullback-leiber divergences.In conference on Learning Thoery(COLT),2011}\\
\vspace{3ex}
\text{P.popescu,S.dragomir,E.slusanschi,N.stanasilla Bounds for Kullback-Leibler Divergence}\\
\text{Electornic Journal of Differential Equations,Volume 2016 (2016),No. 237, pp.1-6}\\
\vspace{3ex}
\text{ T. L. Lai and H. Robbins. Asymptotically efficient adaptive allocation rules.}\\
\text{ Advances in Applied Mathematics,6:4–22, 1985}\\

\end{flushleft}
\end{document}